\documentclass[sigconf]{acmart}

\AtBeginDocument{%
  \providecommand\BibTeX{{%
    \normalfont B\kern-0.5em{\scshape i\kern-0.25em b}\kern-0.8em\TeX}}}

\renewcommand\footnotetextcopyrightpermission[1]{}

\usepackage{ragged2e} 
\usepackage{booktabs,makecell, multirow, tabularx}
\usepackage{algorithm}
\usepackage{algorithmic}
\usepackage{makecell}
\usepackage{graphicx}

\usepackage{amssymb}
\usepackage{amsmath}
\usepackage{units}
\begin{document}

\title{CLIP3D-AD: Extending CLIP for 3D Few-Shot Anomaly Detection with Multi-View Images Generation}

\author{Zuo Zuo}
%\email{nostalgiaz@stu.xjtu.edu.cn}
\affiliation{%
   \institution{National Key Laboratory of Human-Machine Hybrid Augmented Intelligence and Institute of Artificial Intelligence and Robotics, Xi'an Jiaotong University}
   \streetaddress{P.O. Box 1212}
  \city{Xi'an}
  \state{Shanxi}
  \country{China}
 }
\author{Jiahao Dong}
%\email{webmaster@marysville-ohio.com}
\affiliation{%
   \institution{Guangdong Laboratory of Artificial Intelligence and Digital Economy (SZ)}
   \streetaddress{P.O. Box 1212}
  \city{Shenzhen}
  \state{Guangdong}
  \country{China}
 }
 \author{Yao Wu}
%\email{webmaster@marysville-ohio.com}
\affiliation{%
   \institution{Xiamen University}
   \streetaddress{P.O. Box 1212}
  \state{Xiamen}
  \city{Fujian}
  \country{China}
   \postcode{43017-6221}
 }
 \author{Yanyun Qu}
%\email{webmaster@marysville-ohio.com}
\affiliation{%
   \institution{Xiamen University}
   \streetaddress{P.O. Box 1212}
    \state{Xiamen}
  \city{Fujian}
  \country{China}
   \postcode{43017-6221}
 }
 \author{Zongze Wu}
 \authornote{Corrosponding Author.}
%\email{webmaster@marysville-ohio.com}
\affiliation{%
   \institution{Shenzhen University}
   \institution{ Xi'an Jiaotong University}
   \institution{Guangdong Laboratory of Artificial Intelligence and Digital Economy (SZ)}
   \streetaddress{P.O. Box 1212}
  \city{Shenzhen}
  \state{Guangdong}
  \country{China}
   \postcode{43017-6221}
 }

%\author{Anonymous Authors}
%% You do not have to enter your paper ID


\begin{abstract}
   Few-shot anomaly detection methods can effectively address data collecting difficulty in industrial scenarios. Compared to 2D few-shot anomaly detection (2D-FSAD), 3D few-shot anomaly detection (3D-FSAD) is still an unexplored but essential task. In this paper, we propose CLIP3D-AD, an efficient 3D-FSAD method extended on CLIP. We successfully transfer strong generalization ability of CLIP into 3D-FSAD. Specifically, we synthesize anomalous images on given normal images as sample pairs to adapt CLIP for 3D anomaly classification and segmentation. For classification, we introduce an image adapter and a text adapter to fine-tune global visual features and text features. Meanwhile, we propose a coarse-to-fine decoder to fuse and facilitate intermediate multi-layer visual representations of CLIP. To benefit from geometry information of point cloud and eliminate modality and data discrepancy when processed by CLIP, we project and render point cloud to multi-view normal and anomalous images. Then we design multi-view fusion module to fuse features of multi-view images extracted by CLIP which are used to facilitate visual representations for further enhancing vision-language correlation. Extensive experiments demonstrate that our method has a competitive performance of 3D few-shot anomaly classification and segmentation on MVTec-3D AD dataset. 
\end{abstract}

%\begin{CCSXML}
%<ccs2012>
%       <concept_id>10010147.10010178.10010224.10010225</concept_id>
%       <concept_desc>Computing methodologies~Computer vision tasks</concept_desc>
%       <concept_significance>500</concept_significance>
 %      </concept>
%\end{CCSXML}

%\ccsdesc[500]{Computing methodologies~Computer vision tasks}




\maketitle

\begin{figure}
\centering
\includegraphics[width=3.33in, keepaspectratio]{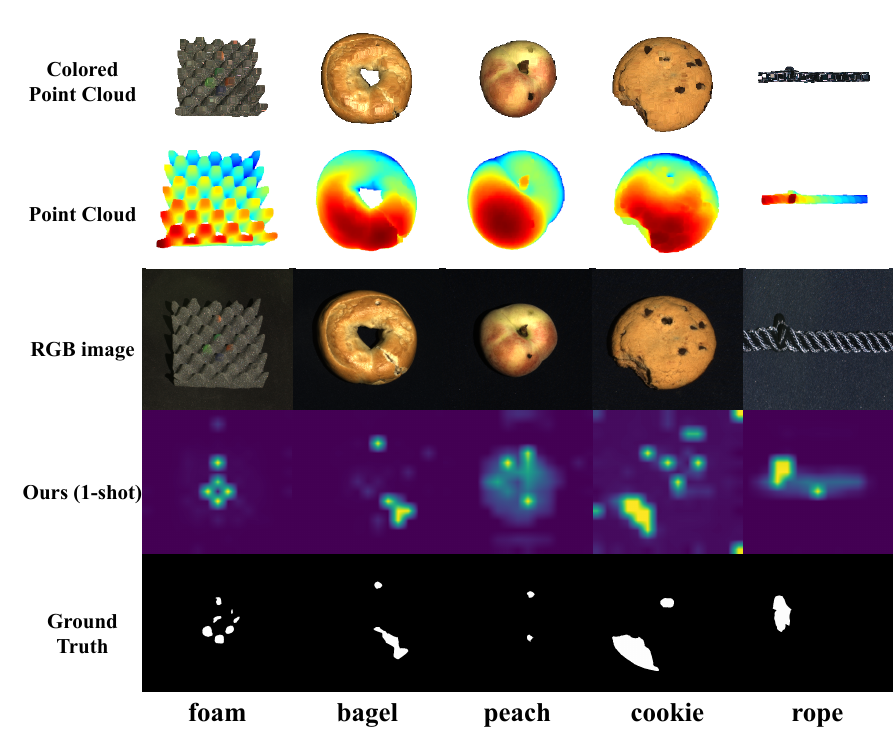}\\
\caption{Examples of 3D anomaly detection on MVTEC-3D AD. The second and third rows are point clouds and RGB images. The fourth row is our prediction of anomaly areas and the last row is ground truth mask.}
\end{figure}

\begin{figure*}
\centering
\includegraphics[width=\textwidth,height=0.6\textwidth]{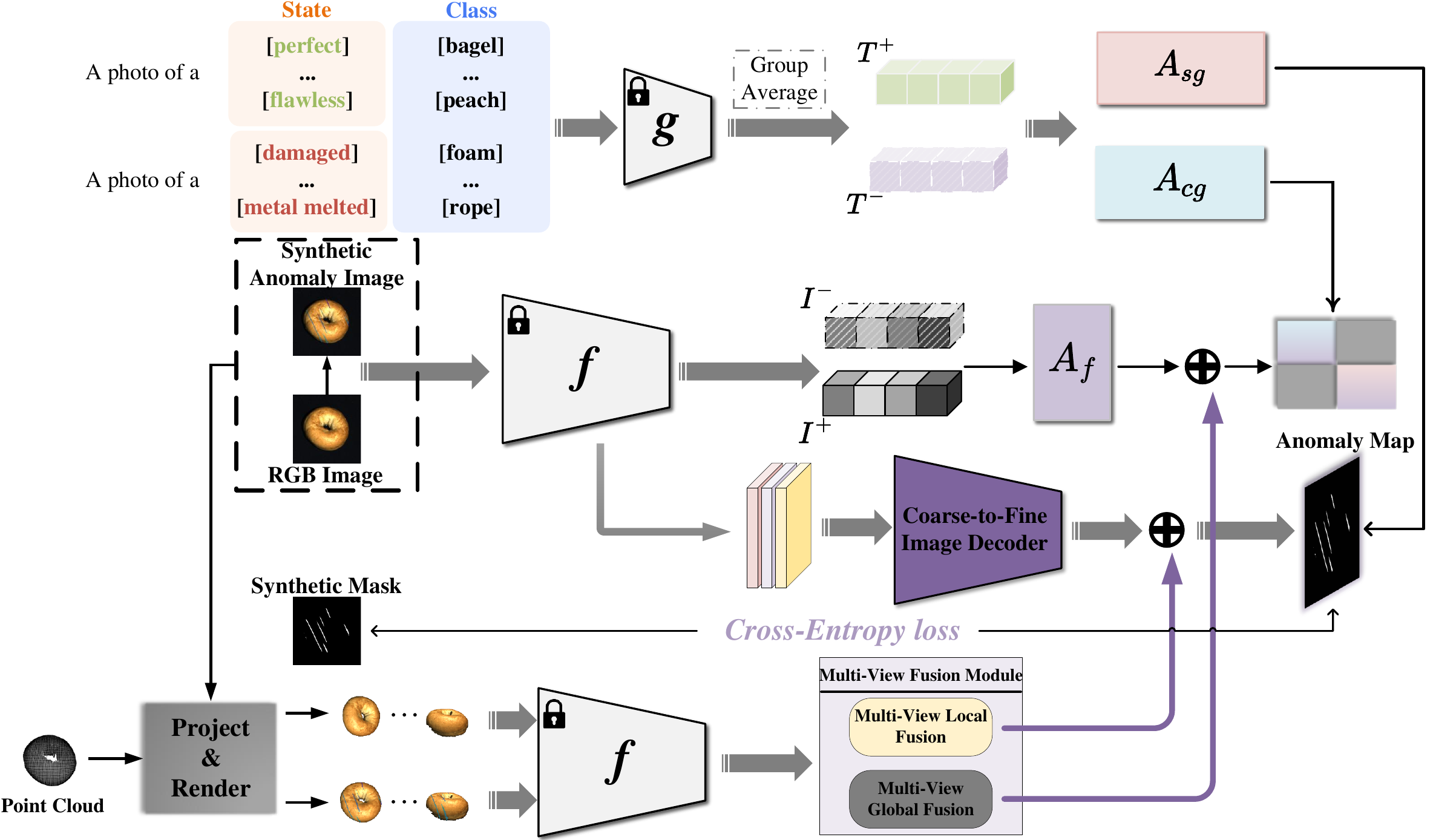}\\
\caption{The framework of CLIP3D-AD. In the training phase, anomaly image is synthesized as negative sample for given normal image as positive sample. We use frozen CLIP image encoder $f(\cdot)$ to extract global and local visual features and frozen CLIP text encoder $g(\cdot)$ to extract normal and anomaly text features. Then we introduce image adapter $A_f(\cdot)$ and two text adapters $A_{cg}(\cdot)$ and $A_{sg}(\cdot)$ to adapt original representations of CLIP. Meanwhile, we project and render point cloud to multi-view images and use multi-view fusion module to fuse multi-view visual features extracted by CLIP. We use fused multi-view features to enhance visual representations. }
\end{figure*}

\section{Introduction}
Industrial anomaly detection (AD) including classification and segmentation aims to finding aberrant areas in given samples \cite{mvtec,rrd} which is addressed by many 2D unsupervised anomaly detection methods. With the increasing demand for anomaly detection task, single-view RGB images are deficient for detecting anomalies on an object~\cite{realiad}. Therefore point cloud with 3D geometry information are utilized in anomaly detection named 3D anomaly detection recently ~\cite{mvtec3d}. These 3D anomaly detection methods use both 2D images and 3D point cloud in a multi-modal mode. Compared to 2D anomaly detection~\cite{dmad,fd}, 3D anomaly detection is not well-explored with limited amount of methods at present \cite{cheatingdepth,ssfa}.

Inspired by 2D anomaly detection methods~\cite{simplenet, prn, dsr}, 3D anomaly detection methods can be categorized in two classes: embedding-based method and student-teacher-based method. Embedding-based methods usually need a memory bank to restore normal features of training samples and detect anomalies by finding feature discrepancy between testing images and memory bank.  Because point cloud pre-trained models are not prevalent in 3D anomaly detection, some student-teacher based methods need to train a teacher model from scratch for point cloud with large amount of data~\cite{3d-st}. Data collecting is a hurdle for so long time in anomaly detection. Lots of 2D few-shot anomaly detection (2D-FSAD) methods emerge recently and they only need a few normal training samples, which reduce tremendously training cost and are easy to deploy. Several 2D-FSAD methods built upon vision-language models such as CLIP~\cite{flant5, pandagpt} have an unexpected performance which are even competitive to many full-shot methods~\cite{promptad, anomalygpt}. But to our knowledge, there is no few-shot methods in 3D anomaly detection. Therefore, a natural question has been raised: Can we leverage the strong representation ability of CLIP in 3D few-shot anomaly detection (3D-FSAD) ? 

However, CLIP trained on a dataset consisted of numerous image-text pairs is not directly geared toward point cloud~\cite{pointclipv2}. Utilizing CLIP to process point cloud has suboptimal performance due to modality gap between images and point cloud. Meanwhile, data used to train CLIP has large distribution discrepancy compared to industrial images. We try to project point cloud to 2D images directly. But directly projecting point cloud into 2D space will lead to loss of geometry information substantially~\cite{mvc, pointclip}. We propose an intriguing idea : find anomalies from different views.

To holistically address the issues mentioned above, we propose a 3D-FSAD method dubbed CLIP3D-AD based on CLIP without memory bank and an exhaustive set of training samples. Specifically, we construct sample pair consisting of normal training image and synthetic anomaly image to fine-tune CLIP for 3D anomaly detection including classification and segmentation. For classification, we introduce an image adapter and class text adapter to transfer global visual features and text features of CLIP for better task-specific correlation. Additionally, we propose a coarse-to-fine image decoder to aggregate and decode intermediate multi-layer visual features of CLIP's image encoder and a segmentation text adapter is used to obtain vision-compatible text embeddings for local vision-language matching. As complement of 2D RGB images, point cloud is introduced in 3D anomaly detection. But directly processing point cloud by CLIP is hindered by modality gap. To make full use of point cloud, we project and render point cloud into multi-view images which eliminates modality gap and enriches the visual information. We design multi-view fusion module to fuse global and local visual features extracted from multi-view images by CLIP. Then these aggregated features are added to adapted features of RGB image to facilitate visual representation. During the whole fine-tuning phase, image and text encoders of CLIP are frozen and the proposed modules are optimized by combination of bidirectional contrastive and cross-entropy loss. To sum up, our main contributions are as follows :
\begin{itemize}
\item  As far as we know, we are the first to transfer few-shot anomaly detection with vision-language models into 3D few-shot anomaly detection.  

\item  We propose to adapt CLIP for 3D few-shot anomaly detection with image and text adapters for anomaly classification and coarse-to-fine image decoder for anomaly segmentation.

\item  We propose the concept of "Find Anomalies From Different Views" that is to project and render point cloud into multi-view images to enhance visual representations and alleviate domain gap between image and point cloud processed by CLIP.

\item  Experimental results demonstrate superior performance of
CLIP3D-AD in both 3D few-shot anomaly classification and segmentation.
\end{itemize}

\begin{figure}
\centering
\includegraphics[width=3.33in, keepaspectratio]{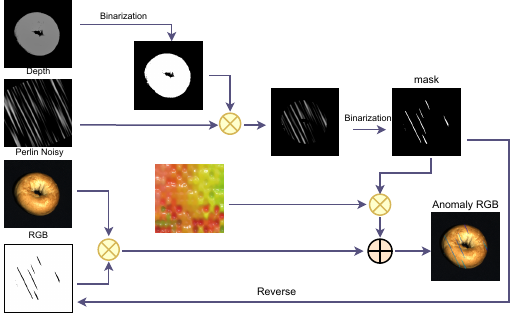}\\
\caption{Pipeline of anomalies generation.}
\end{figure}

\section{Related Work}
\subsection{3D Anomaly Detection}
Since the research on 2D image anomaly detection has been quite in-depth~\cite{msflow, patchcore, rd, sspc,bds} , 3D anomaly detection methods are not yet mature. MVTec3D-AD dataset \cite{mvtec3d} is the first public 3D industrial anomaly detection dataset, in which point cloud can provide geometric position information for anomaly detection. Then, how to utilize the information provided by point clouds effectively is a crucial issue for 3D anomaly detection.
In previous works, Shape-Guided~\cite{shape-guide} proposed two expert networks, which extract 3D shape and RGB information to construct shape-guided memory banks to detect shape and appearance anomalies. M3DM \cite{m3dm} aligns the 3D feature to 2D space and construct RGB, 3D and fusion memory banks. Constructing memory bank will lead to large memory usage and time cost. To reduce the cost of anomaly detection, EasyNet \cite{easynet} proposed a simple reconstruction network with a high inference speed which contains an information entropy module based on attention mechanism. \cite{cfm} proposes a fast framework that learns to map features from one modality to the other on nominal samples and detects anomalies by inconsistencies between observed and mapped features. In order to make full use of the 3D information provided by point cloud, we generate multi-view images from point clouds for 3D anomaly detection.

\subsection{Few-shot Anomaly Detection}
Due to the difficulty in collecting samples in industrial scenarios, it often cannot meet the data requirements for training. Therefore, few-shot anomaly detection \cite{coft-ad, fsfa, regad} has been developed to meet the demand of this situation. FastRecon \cite{fastrecon} reconstructs normal samples based on features extracted from few normal samples by learning linear transformation. GraphCore \cite{graphcore} proposes to extract visual isometric invariant features using GNN and constructs a memory bank for few-shot anomaly detection. As the development of vision-language models, it becomes possible to detect anomalies through language guidance. Recently, several methods transfer strong representation ability of large multimodal (i.e. vision and language) models into few-shot anomaly detection. WinCLIP \cite{winclip} is the first work to explore the language-guided zero-shot anomaly classification and segmentation which introduces compositional prompt ensemble and extracts dense visual features of different scales through window based CLIP. Myriad~\cite{myriad} adopts MiniGPT-4 \cite{minigpt4} and introduces a domain adapter and a vision expert instructor for industrial anomaly detection.
Our work is extended on CLIP. With projecting point cloud to multi-view images, our method mitigates domain gap by fine-tuning the text and visual representations extracted by CLIP and achieves better performance in 3D-FSAD.

\begin{figure}
\centering
\includegraphics[width=3.33in, keepaspectratio]{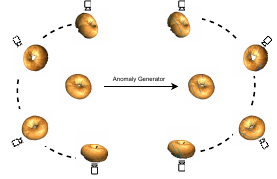}\\
\caption{Example of generated multi-view images.}
\end{figure}

\section{Method}
\subsection{Revisit of CLIP}
Contrastive language image pre-training (CLIP) \cite{clip} is one of the most commonly used vision-language models. The goal of CLIP is to predict the correct pairings of (image, text) examples. Trained on 400 million image-text pairs collected from the web, CLIP has so strong generalization ability that it can be successfully transferred in many downstream tasks. It has an image encoder $f(\cdot)$ and a text encoder $g(\cdot)$ which extract visual and language representations for learning. Benefiting from remarkable representative ability of CLIP, WinCLIP~\cite{winclip} explores CLIP in 2D zero/few-shot anomaly detection and reaches a satisfactory result. 

However, CLIP has been under explored in 3D anomaly detection. We extend it with coarse-to-fine image decoder and adapters in 3D few-shot anomaly detection. Meanwhile we take advantage of geometry information of point cloud to enhance visual representations for better image-language matching.

\subsection{Overview of CLIP3D-AD}
To explore CLIP in 3D-FSAD, we propose CLIP3D-AD to extend CLIP in 3D-FSAD with adapters \cite{clipadapter, tipadapter} and coarse-to-fine image decoder. Besides, we project point cloud into multi-view images to facilitate visual representation and avoid processing point cloud directly using CLIP. The overall framework of CLIP3D-AD is illustrated in Figure 2. Given normal training image $x^+\in \mathcal X_{train}$ and its corresponding point cloud $p^+$, we firstly synthesize anomalous image $x^-\in $ as negative sample and generate mask simultaneously. We use rotation matrices to rotate $p^+$ into different point cloud and utilize $x^+$ and $x^-$ to render them to obtain normal and anomalous multi-view images as $\boldsymbol V^{+}_{1:v}$ and $\boldsymbol V^{-}_{1:v}$. Based on CLIP, we feed $x^+$ and $x^-$ into frozen image encoder $f(\cdot)$ to extract [CLS] tokens and features of specific layers. We introduce image adapter and coarse-to-fine image decoder to fine-tune and decode visual representation from CLIP. Then we use text adapters to transfer text embeddings for better correlation between vision and language. Meanwhile, to benefit from geometry information of point cloud, we design multi-view fusion module to aggregate visual representations encoded by CLIP from rendered images which are used to enhance visual embeddings of $x^+$ and $x^-$.

\subsection{CLIP3D-AD}
\subsubsection{Anomaly simulation}
 To train our proposed modules, we generate anomaly counterpart $x^-$ for given normal training image $x^+$ as depicted in Figure 3. Following EasyNet~\cite{easynet} and DRAEM~\cite{draem} we firstly generate a foreground mask using depth image, and generate random Perlin noisy. Then we apply this foreground mask on Perlin noisy to obtain a Perlin noisy map. The binarization of noisy map is mask map which is used to apply on randomly selected images to generate anomalies. Anomaly regions are decided by the reverse of mask map. Finally we add synthetic anomalies on anomaly regions of original images to obtain RGB images with anomalies $x^-$ and ground truth masks $\hat{m}$.

\begin{figure}
\centering
\includegraphics[width=3.33in, keepaspectratio]{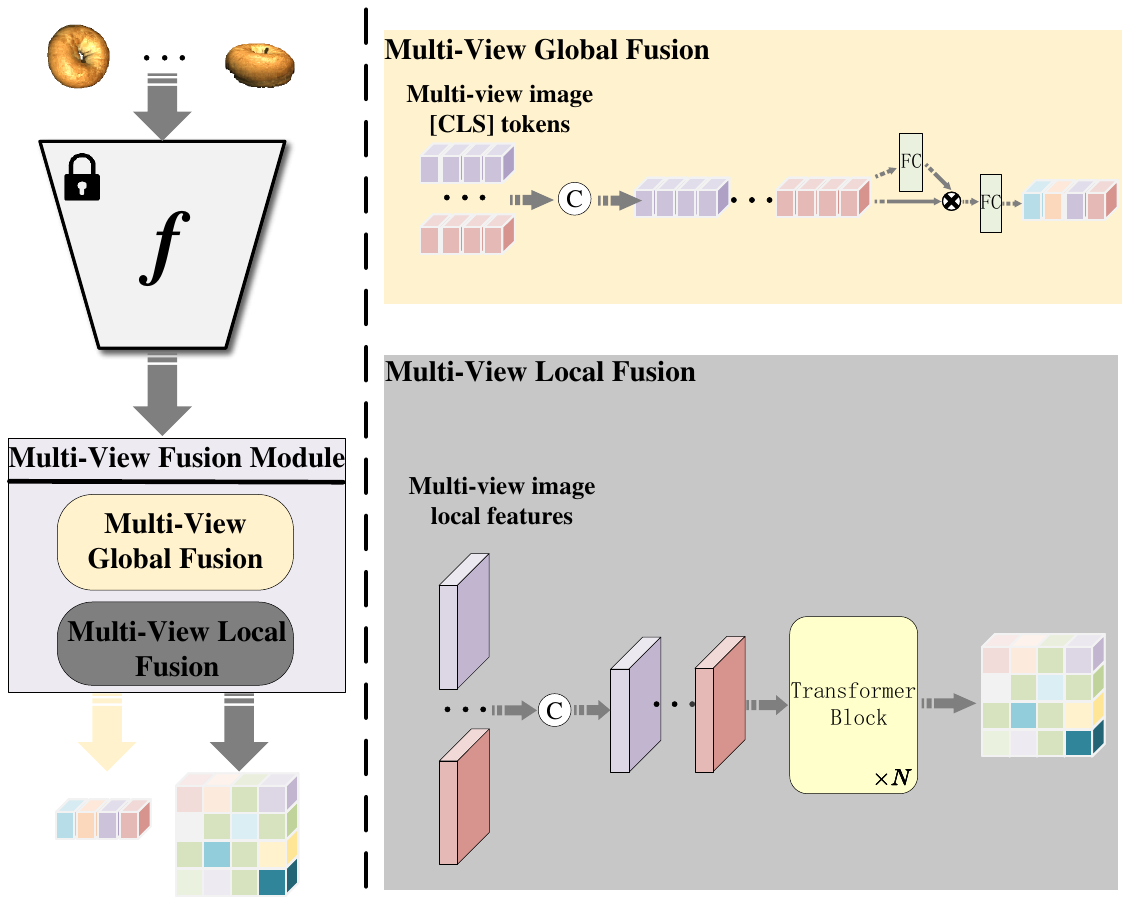}\\
\caption{Architecture of multi-view fusion module consisting of multi-view global fusion and multi-view local fusion.}
\end{figure}

\subsubsection{Multi-View Images Generation}
3D anomaly detection involves two modalities processing: 2D images and 3D point cloud. 3D point cloud has sparsity and irregularity and is different from 2D images, which lead to large domain discrepancy when processed by CLIP. We address this modality discrepancy by making an insight that we project point cloud into multi-view images using geometry transformation and rendering operation~\cite{cpmf}. As shown in Figure 4, we use original normal image to render its corresponding point cloud $P_{3D}$ to get multi-view images. Meanwhile, we obtain multi-view anomaly images by rendering  $P_{3D}$ with synthetic anomaly images. To see object from different views, we set a series of various rotation matrices 
\begin{math} \{ \boldsymbol R_1,\boldsymbol R_2,\dots,\boldsymbol R_v \},\boldsymbol R\in \mathbb{R}^{3 \times 3}\end{math}. Different $\boldsymbol R$ indicates rotation at different angles $\theta_x,\theta_y,\theta_z$ along $x,y,z$ axes, which is defined as
\begin{equation}
\begin{aligned}
  \boldsymbol R =  &\left[
 \begin{array}{ccc}
     cos \,\theta_x & sin \,\theta_x& 0 \\
     -sin\, \theta_x& cos\, \theta_x & 0 \\
     0 & 0 & 1 
 \end{array}
 \right]     \left[
 \begin{array}{ccc}
     cos \,\theta_y & 0& sin \,\theta_y \\
     0& 1 & 0 \\
     -sin\, \theta_y & 0 & cos \,\theta_y 
 \end{array}
 \right]  \\
 &\left[ \begin{array}{ccc}
     1 & 0 & 0              \\
     0 &cos\, \theta_z & sin\, \theta_z \\
     0 &-sin\, \theta_z& cos \,\theta_z  \\
       
 \end{array}
 \right]
 \end{aligned}
\end{equation}
The $k$-th rotated point cloud of $P_{3D}$ is calculated as
\begin{equation}
\begin{aligned}
P_{3D,k} = \boldsymbol R_kP_{3D}.
 \end{aligned}
\end{equation}

After rotating $P_{3D}$, we use camera model parameter $U$ and  render function $\psi$ to obtain rendered images $\boldsymbol V_k$ and 2D position $P_{2D}$:
\begin{equation}
\begin{aligned}
\boldsymbol V_k = \psi(P_{3D,k}, U)
 \end{aligned}
\end{equation}
\begin{equation}
\begin{aligned}
[P_{2D},1]^T = \frac{1}{Z_c}KT[P_{3D,k},1]^T,
\end{aligned}
\end{equation}
where $K$ and $T$ are the intrinsic and extrinsic matrix of the 
rendered camera. $Z_c$ is the normalized 
coordinate coefficient.
Generally for point cloud $P_{3D}$, it is rotated by series of rotation matrices \begin{math} \{ \boldsymbol R_1,\boldsymbol R_2,\dots,\boldsymbol R_v \}\end{math} and is rendered to generate two sets of 2D multi-view images \begin{math} \{ \boldsymbol V_1^+,\boldsymbol V_2^+,\dots,\boldsymbol V_v^+ \}\end{math} and \begin{math} \{ \boldsymbol V_1^-,\boldsymbol V_2^-,\dots,\boldsymbol V_v^- \}\end{math} by $x^+$ and $x^-$, where $v=27$.

\begin{table*}[h]

\caption{Anomaly detection performance evaluated by the metric of I-AUROC on the MVTEC-3D AD dataset. Red indicates the best result, and blue displays the second-best.}
\resizebox{\textwidth}{!}{%
\begin{tabular}{cl|cccccccccc|c}
\toprule
\multicolumn{2}{c|}{Method}& bagel & cable\_gland & carrot & cookie & dowel & foam & peach & potato & rope & tire & mean \\ \hline
{\color[HTML]{333333} }& AST \cite{ast} & 70.7 & 42.2 & 54.8 & 49.0 & 53.8 & 46.4 & 51.9 & 49.7 & 72.0 & 41.9 & 53.2 \\
{\color[HTML]{333333} }& Patchcore+FPFH \cite{btf} & 62.2  & 53.4 & 54.0 & 55.9 & 54.7 & 63.3 & 49.6 & 60.7 & 88.8 & 56.6 & 59.9 \\
\multirow{5}{*}[+5ex]{1-shot}
{\color[HTML]{333333} }& EasyNet \cite{easynet} & 61.4 & 21.2 & 52.0 & 75.9 & 56.5  & 62.8 & 65.7 & 63.0 & 94.6 & 47.7 & 60.1 \\
{\color[HTML]{333333} }& ShapeGuided \cite{shape-guide} & 65.9  & 44.4 & 62.3 & 93.8 & 59.3 & 57.6 & 67.6 & 42.8 & 93.3 & 62.9 & \textcolor{blue}{65.0} \\
{\color[HTML]{333333} } & Ours & 91.1 & 74.1 & 79.7 & 60.0 & 56.3  & 75.7 & 85.7 & 61.3 & 91.5 & 57.4 & \textcolor{red}{73.3} \\ \hline
& AST \cite{ast} & 71.9  & 43.4 & 54.5 & 50.8 & 53.7 & 46.1 & 51.6 & 50.4 & 75.8 & 40.2 & 53.8 \\
& Patchcore+FPFH \cite{btf} & 63.2  & 47.7 & 55.4 & 64.5 & 58.3 & 61.1 & 55.0 & 56.6 & 88.2 & 64.4 & 61.4 \\
\multirow{5}{*}[+5ex]{2-shot}
& EasyNet \cite{easynet} & 47.6 & 76.1 & 52.6 & 60.2 & 31.7 & 52.3 & 71.9 & 76.1  & 61.2 & 51.2 & 58.1 \\
& ShapeGuided \cite{shape-guide} & 47.9 &46.0 & 60.5 & 95.9 & 55.3 & 50.2 & 69.7 & 41.3 & 93.6 & 79.3 & \textcolor{blue}{64.0} \\
& Ours & 83.5 & 79.0 & 82.7 & 50.9 & 58.8 &  88.6 & 72.5 &  63.2 & 90.9  & 70.3 &  \textcolor{red}{74.0}    \\ \hline
& AST \cite{ast}& 70.1 & 42.9 & 55.7 & 51.8 & 54.0 & 46.6 & 52.0 & 49.8  & 72.6 & 39.8 & 53.5 \\
& Patchcore+FPFH \cite{btf}& 52.3 & 54.0 & 59.2 & 62.9 & 59.4 & 58.2 & 61.6 & 70.5  & 91.8 & 73.5 & 64.3 \\
\multirow{5}{*}[+5ex]{4-shot}
& EasyNet \cite{easynet} & 67.5 & 36.3 & 54.7 & 69.1 & 72.4 & 50.2 & 74.3 & 63.1 & 44.9 & 53.7 & 58.6 \\
& ShapeGuided \cite{shape-guide} & 65.4 & 48.8 & 73.1 & 96.5 & 69.8 & 59.1 & 68.1  & 49.9 & 92.2 & 75.1 & \textcolor{blue}{69.8} \\
& Ours & 89.9 & 68.0 & 74.1 & 57.7 & 65.1 & 81.6 &  79.2 &66.6  & 67.5& 57.4  &  \textcolor{red}{70.7}    \\ \bottomrule
\end{tabular}%
}
\end{table*}

\subsubsection{Adapters And Coarse-To-Fine Image Decoder}
Given image pair $x^+, x^-\in \mathbb R^{H\times W\times 3}$, where H, W denote the height and width, [CLS] tokens encoded by CLIP are denoted as $I^+, I^-\in \mathbb R^{1\times C}$, where $C$ is dimension of [CLS] tokens. The feature from $l\in L$ stage of CLIP is denoted as $\mathcal F_l \in \mathbb R^{N_p\times D}$, where $N_p$ is the number of tokens and $D$ is feature dimension. We use CLIP text encoder $g(\cdot)$ to encode two sets of text prompts and average them respectively to obtain normal text feature and anomaly text feature depicted as $T^+$ and $T^-$. To mitigate distribution discrepancy between pre-trained data and target data, we add adapters and decoder for task-oriented features. For global classification, we introduce image adapter denoted as $A_f(\cdot)$ and classification text adapter denoted as $A_{cg}(\cdot)$. Adapted classification text features and visual representations are defined as:
\begin{equation}
T^+_c = A_{cg}(T^+), \;T^-_c = A_{cg}(T^-),\; T^+_c,\,T^-_c\in \mathbb R^{1\times C}
\end{equation}
\begin{equation}
I^+_A = A_f(I^+), \;I^-_A = A_f(I^-),\; I^+_A,\,I^-_A\in \mathbb R^{1\times C}.
\end{equation}

For local segmentation, we introduce segmentation text adapter denoted as $A_{sg}(\cdot)$ and coarse-to-fine image decoder consisting of transformer blocks denoted as $C2F\_Decoder(\cdot)$. Adapted segmentation text feature $T_s$ is calculated as follows:
\begin{equation}
T_s = Concat\left[A_{sg}(T^+), A_{sg}(T^-)\right],\;T_s \in \mathbb R^{2\times C}.
\end{equation}
We choose features of three different stages from CLIP and concatenate them as local representation. For better matching text and local image features, we design coarse-to-fine image decoder to transfer original visual features to language-compatible local features as follows:
\begin{equation}
\mathcal F = C2F\_Decoder(Concat(\mathcal F_l|l\in L)),\, \mathcal F \in \mathbb R^{N_p\times C},
\end{equation}
 After adaptation, we calculate the similarity between normalized language-compatible local features $\mathcal F'=\frac{\mathcal F}{\Vert \mathcal F \Vert}$ and normalized adapted segmentation text features $T_s'=\frac{T_s}{\Vert T_s \Vert}$ as 
\begin{equation}
 M_{sim} = \nicefrac{\mathcal F' \cdot T_s'^T}{\gamma} , \; M_{sim} \in \mathbb R^{N_p\times 2},
\end{equation}
where $\gamma=0.07$. Then we calculate softmax of $M_{sim}$ to obtain anomaly map as:
\begin{equation}
 S_{map} = Softmax(M_{sim})[:,i], S_{map}\in \mathbb R^{N_p\times 1},
\end{equation}
where $i=2$. To align with original solution of input image, we interpolate and reshape anomaly map to $S_{map}\in \mathbb R^{H\times W\times 1}$.

\subsubsection{Multi-View Fusion Module}
After generating a series of multi-view images, we select 5 views of images \begin{math} \{\boldsymbol V_i^{+/-}|i\in \left[5, 9, 14, 19, 27\right]\}\end{math} which almost cover overall perspective. We design multi-view fusion module (MVFM) as illustrated in Figure 5 which consists of global fusion and local fusion modules. Global fusion module consists of channel attention~\cite{senet} which assigns weights to features of different views and a fully connected layer for global fusion. Local fusion module is composed of two transformer blocks and a final projected layer for local fusion. Global fusion module aggregates all [CLS] tokens encoded by CLIP and local fusion module aggregates all features extracted from last stage of CLIP image encoder. After aggregation, we add global fused and local fused features on corresponding adapted global features and adapted local features to enhance the visual representation which is more discriminative and language-compatible.

\begin{table*}[h]
\caption{Anomaly localization performance measured by the metric of AUPRO on the MVTEC-3D AD dataset. Red indicates the best result, and blue displays the second-best.}
\resizebox{\textwidth}{!}{%
\begin{tabular}{cl|cccccccccc|c}
\toprule
\multicolumn{2}{c|}{Method}& bagel & cable\_gland & carrot & cookie & dowel & foam & peach & potato & rope & tire & mean \\ \hline
{\color[HTML]{333333} }& AST \cite{ast} & 75.9 & 73.3 & 88.0 & 60.2 & 79.4 & 44.0 & 84.0 & 85.9 & 75.8 & 74.0 & 74.0 \\
{\color[HTML]{333333} }& Patchcore+FPFH \cite{btf} & 92.8 & 76.8 & 96.7 & 92.8 & 84.6 & 71.9 & 95.9 & 96.1 & 90.8 & 84.1 & \textcolor{blue}{88.3} \\
\multirow{5}{*}[+5ex]{1-shot}
{\color[HTML]{333333} }& EasyNet \cite{easynet} & 79.6 & 75.1 & 91.0 & 69.8 & 85.8 & 49.4 & 69.0 & 88.1 & 71.8 & 75.4 & 75.5 \\
{\color[HTML]{333333} }& ShapeGuided \cite{shape-guide} & 95.9 & 71.6 &93.5 & 94.1 & 86.4 & 63.8 & 94.0 & 96.3 & 88.8 & 90.1 & 87.4 \\
{\color[HTML]{333333} } & Ours & 97.0 & 85.6 & 93.8 & 85.5 & 87.1 & 79.7 & 91.8 & 95.0 &88.6  & 88.9 & \textcolor{red}{89.3} \\ \hline
& AST \cite{ast} & 75.9 & 74.0 & 87.8 & 62.2 & 79.5 & 43.6 & 83.7 & 85.6 & 76.5 & 72.8 & 74.2 \\
& Patchcore+FPFH \cite{btf} & 94.6 & 76.5 & 96.7 & 93.4 & 85.0 & 67.4 & 96.2 & 96.6 & 91.0 & 88.3 & 88.6 \\
\multirow{5}{*}[+5ex]{2-shot}
& EasyNet \cite{easynet} & 77.8 & 62.9 & 92.6 & 59.1 & 58.8 & 57.2 & 21.1 & 15.2 & 43.1 & 4.7 & 49.3 \\
& ShapeGuided \cite{shape-guide} & 96.6  & 73.2 & 96.5 & 95.5 & 86.5 & 71.4 & 95.3  & 96.3  & 89.3 & 91.3 & \textcolor{blue}{89.2} \\
& Ours &  95.5 & 92.7& 95.5 & 85.1 &  84.3 &  84.9 & 91.6  &  95.4  &  89.2 & 87.4  & \textcolor{red}{90.2}   \\ \hline
& AST \cite{ast} & 74.7 & 73.7 & 87.7 & 61.3 & 79.6 & 41.3 & 84.3 & 85.9 & 75.9 & 74.2 & 73.9 \\
& Patchcore+FPFH \cite{btf}& 95.7 & 79.7 & 97.2 & 95.1 & 87.5 & 75.3 & 96.5 & 97.3 & 91.2 & 88.0 & \textcolor{blue}{90.4} \\
\multirow{5}{*}[+5ex]{4-shot}
& EasyNet \cite{easynet} & 70.7 & 13.8 & 86.9 & 72.0 & 39.3 & 53.6 & 86.2 & 90.2 & 15.5 & 66.3 & 56.5 \\
& ShapeGuided \cite{shape-guide} & 97.3 & 78.9 & 97.3 & 95.4 & 90.4 & 83.6 & 95.7 & 97.5 & 89.6 & 92.1 & \textcolor{red}{91.8} \\
& Ours &  95.2  &  91.6  &  94.1 &  86.0&  88.5 &82.3  &  94.4  & 95.1 &  71.5   & 90.1   & 88.9     \\ \bottomrule
\end{tabular}%
}
\end{table*}

\subsubsection{Loss Function}
To optimize our proposed modules, we use bidirectional contrastive loss $\mathcal{L}_{con}$ as classification loss and binary cross-entropy loss as segmentation loss. Bidirectional contrastive loss $\mathcal{L}_{con}$ consists of image-to-text contrastive loss $\mathcal{L}_{i2t}$
and text-to-image contrastive loss $\mathcal{L}_{t2i}$ \cite{moco} calculated as 
\begin{equation}
\begin{aligned}
\mathcal{L}_{i2t} = -log\frac{exp(s(I^+_A, T^+_c))}{exp(s(I^+_A, T^+_c))+exp(s(I^+_A, T^-_c))}\\
-log\frac{exp(s(I^-_A, T^-_c))}{exp(s(I^-_A, T^-_c))+exp(s(I^-_A, T^+_c))},
\end{aligned}
\end{equation}
\begin{equation}
\begin{aligned}
\mathcal{L}_{t2i} = -log\frac{exp(s(I^+_A, T^+_c))}{exp(s(I^+_A, T^+_c))+exp(s(I^-_A, T^+_c))}\\
-log\frac{exp(s(I^-_A, T^-_c))}{exp(s(I^-_A, T^-_c))+exp(s(I^+_A, T^-_c))},
\end{aligned}
\end{equation}
where $s(\cdot,\cdot)$ is consine similarity function. Bidirectional contrastive loss $\mathcal{L}_{con}$ is defined as 
\begin{equation}
    \mathcal{L}_{con} = \frac{1}{2}(\mathcal{L}_{i2t} + \mathcal{L}_{t2i}).
\end{equation}

For segmentation loss, the objective is defined as binary cross-entropy between anomaly map $S_{map}$ and ground truth mask $\hat{m}$:
\begin{equation}
    \mathcal{L}_{seg} = BinaryCrossEntropy(S_{map},\hat{m})
\end{equation}
Ultimately, the total objective is summarized as follows:
\begin{equation}
    \mathcal{L}_{tot} = \mathcal{L}_{seg} + \mathcal{L}_{con}
\end{equation}
\subsection{Anomaly Detection with CLIP3D-AD}
In testing phase, we calculate anomaly map $S_{map}^{test}$ of test image $x_{test}\in \mathcal X_{test}$ followed Equation (9) and (10) which is regarded as segmentation results. Additionally, we calculate anomaly score of $x_{test}$ as classification results. Positive score $S^+$ and negative score $S^-$ are firstly calculated as follows:
\begin{equation}
S^+(x_{test}) = \frac{exp(\frac{s(I_A^{test}, T_c^+)}{\tau})}{exp(\frac{s(I_A^{test}, T_c^-)}{\tau})+exp(\frac{s(I_A^{test}, T_c^+)}{\tau})},
\end{equation}
\begin{equation}
S^-(x_{test}) = \frac{exp(\frac{s(I_A^{test}, T_c^-)}{\tau})}{exp(\frac{s(I_A^{test}, T_c^-)}{\tau})+exp(\frac{s(I_A^{test}, T_c^+)}{\tau})},
\end{equation}
where $\tau=0.07$. Then anomaly score $A_{score}$ which is assigned to testing image for anomaly classification is obtained as
\begin{equation}
A_{score}(X_{test}) = \frac{S^-}{S^-+S^+}+\text{max} \,S_{map}^{test}.
\end{equation}

\section{Experiments}

\subsection{Experiment Setup}
\subsubsection{Dataset}
Our experiments are conducted on MVTec-3D AD dataset \cite{mvtec3d}, the first 3D anomaly detection dataset, which provides RGB images and point cloud . MVTec-3D AD includes 2656 training samples and 1137 testing samples in 10 categories which is the most commonly used 3D anomaly detection dataset.
\subsubsection{Evaluation Metrics}
We evaluate the anomaly classification performance of our method with the Area Under the Receiver Operator Curve (I-AUROC) \cite{pni} and Area Under the Precision-Recall curve (AUPR) \cite{musc}. For segmentation, we report the pixel-wise AUROC (P-AUROC) and the Per-Region Overlap (AUPRO)~\cite{cutpaste} score which is defined as the average relative overlap of the binary prediction with each connected component of the ground truth.
\subsubsection{Implementation details}
We build CLIP3D-AD upon CLIP implemented by OpenCLIP with ViT-B/16+ \cite{openclip, rsl} and load LAION-400M \cite{laion} pre-trained checkpoints. The text prompt we used is CPE proposed in WinCLIP~\cite{winclip}. All input images are resized to 240 $\times$ 240 for accordance with CLIP’s settings. The intermediate layers of the image encoder we choose are 6, 9, 12, that is $L \in \{6, 9, 12\}$. Adam optimizer is used and learning rate is set to 1e-5,5e-5,5e-4,5e-4,1e-4 for classification and segmentation text adapters, image adapter, coarse-to-fine image decoder and multi-view fusion module. We choose 1-shot,2-shot,4-shot as our few-shot setup. All the experiments are performed on NVIDIA RTX 3090 GPUs.

\begin{table}[]
\caption{Additional metrics AUPR and P-AUROC comparison with SOTA approaches on MVTEC-3D AD dataset. Red indicates the best result, and blue displays the second-best.}
\resizebox{\columnwidth}{!}{%
\begin{tabular}{cl|cccc}
\toprule
\multicolumn{1}{l}{k-shot} & Method       & I-AUROC & AUPR   & P-AUROC & AUPRO  \\ \hline
                           & Patchcore+FPFH          & 59.9    & 84.8 & \textcolor{blue}{96.8}    & \textcolor{blue}{88.3} \\
                           & AST          & 53.2    & 81.4 & 91.4    & 74.0 \\
\multirow{5}{*}[+5ex]{1-shot}          & EasyNet      & 60.1    & 84.7 & 91.0    & 75.5 \\
                           & Shape-Guided & \textcolor{blue}{65.0}    & \textcolor{blue}{87.3} & 96.4    & 87.4 \\
                           & Ours         &  \textcolor{red}{73.3$\pm$2.5 }  & \textcolor{red}{91.0$\pm$1.2} &  \textcolor{red}{96.9$\pm$0.4}   &  \textcolor{red}{89.3$\pm$0.9}    \\ \hline
                           & Patchcore+FPFH          & 61.4    & 85.3 & 97.0    & 88.6 \\
                           & AST          & 53.8    & 81.8 & 91.5    & 74.2 \\
\multirow{5}{*}[+5ex]{2-shot}          & EasyNet      & 58.1    & 83.7 & 83.6    & 49.3 \\
                           & Shape-Guided & \textcolor{blue}{64.0}    & \textcolor{blue}{86.5} & 97.0    & \textcolor{blue}{89.2} \\
                           & Ours         &\textcolor{red}{74.0$\pm$2.3}     & \textcolor{red}{90.7$\pm$0.9}  &  \textcolor{red}{97.0$\pm$0.4}   &  \textcolor{red}{90.2$\pm$1.1}    \\ \hline
                           & Patchcore+FPFH          & 64.3    & 86.5 & \textcolor{blue}{97.6 }   & \textcolor{blue}{90.4} \\
                           & AST          & 53.5    & 81.7 & 91.4    & 73.9 \\
\multirow{5}{*}[+5ex]{4-shot}          & EasyNet      & 58.6    & 83.6 & 86.0    & 56.5 \\
                           & Shape-Guided & \textcolor{blue}{69.8}    & \textcolor{blue}{88.9} & \textcolor{red}{97.9}    & \textcolor{red}{91.8} \\
                           & Ours         &\textcolor{red}{70.7$\pm$0.6}    &  \textcolor{red}{89.7$\pm$0.3}  & 96.9$\pm$0.1   &  88.9$\pm$0.5  \\ \bottomrule
\end{tabular}%
}
\end{table}

\subsection{Comparison with State-of-the-Arts}
We compare few-shot anomaly classification and segmentation performance of our method with prior state-of-the-arts works EasyNet~\cite{easynet}, AST~\cite{ast}, Shape-guided~\cite{shape-guide} and Patchcore+FPFH~\cite{btf} in 1-shot,2-shot and 4-shot setup. Table 1 shows the classification results I-AUROC of each subclass in MVTEC-3D AD and it is demonstrated that average results of CLIP3D-AD reach 73.3\%,74.0\% and 70.7\% for 1-shot,2-shot and 4-shot which outperform other methods by a wide margin.  In particular, our I-AUROC is 8.3\%, 10\%, and 0.9\% higher than the second best one in 1-shot, 2-shot and 4-shot respectively. The segmentation results AUPRO of each class are shown in Table 2. We improve AUPRO upon Patchcore+FPFH~\cite{btf} by 1\% in 1-shot and upon Shape-guided~\cite{shape-guide} by 1\% in 2-shot. Our AUPRO drops in 4-shot and we analyse that synthetic anomalies have low quality during 4-shot training phase. To further validate the effectiveness of CLIP3D-AD, we present two extra metrics AUPR and P-AUROC for classification and segmentation as illustrated in Table 3. It is notable that though segmentation results of Shape-guided and Patchcore+FPFH are close to ours and Shape-guided even outperforms ours in 4-shot, they both save normal features as memory bank. Size of the memory used by Patchcore+FPFH is even more than 200 MB reported in EasyNet which leads to large memory usage and hurdles real-world application. Due to randomness in anomalies generation, we 
conduct experiments with five different seeds and report standard deviation of used metrics in Table 3.

\subsection{Comparison with Full-Shot Methods}
Full-shot methods mean that other 3D anomaly detection methods which use all training samples compared to few-shot setup. We compare CLIP3D-AD with other full-shot 3D anomaly detection methods. As illustrated in Table 4, performance of CLIP3D-AD outperforms most of methods such as EasyNet~\cite{easynet}, 3D-ST~\cite{3d-st} on segmentation metric AUPRO and achieves competitive results on classification metric I-AUROC though we use fewer training samples. Notably, AUPRO of proposed CLIP3D-AD outperforms all other full-shot methods attributing to strong representative ability of CLIP and our adaptation and multi-view enhancement.

\begin{table}[]
    \caption{Comparison of I-AUROC and AUPRO between MV-3DADCLIP
and full-shot methods on MVTEC-3D AD dataset.}
    \begin{tabular}{cccc}
       \toprule
        Method  & K-shot & I-AUROC & AUPRO \\
        \hline
        Ours  & 1-shot   & 73.3  & 89.3 \\
        Ours  & 2-shot   &  74.0 & 90.2 \\
        Ours  & 4-shot   &  70.7 & 88.9 \\
        \midrule
        Depth AE \cite{mvtec3d} & full-shot  & 59.5 & 48.1\\
        Depth VM \cite{mvtec3d} & full-shot  & 55.5 & 33.5 \\  
        Voxel GAN \cite{mvtec3d} & full-shot  & 51.7 & 63.9 \\ 
        Voxel AE \cite{mvtec3d} & full-shot  & 53.8 & 56.4 \\  
        Voxel VM \cite{mvtec3d} & full-shot  & 60.9 & 47.1 \\ 
        3D-ST \cite{3d-st} & full-shot     & 83.3 & 83.3 \\ 
        AST \cite{ast} & full-shot & 93.7 & - \\ 
        EasyNet \cite{easynet} & full-shot & 92.6 & 82.1 \\ 
        \bottomrule
    \end{tabular}
    \label{tab:plain}
\end{table}

\begin{table}[t]
\caption{Ablation studies on effectiveness of multi-view images. (Classification: I-AUROC/AUPR. Segmentation: P-AUROC/AUPRO)}
  \centering
    \begin{tabular}{cccc}
    \toprule
       Method  & K-shot & Classification  & Segmentation \\
    \midrule
     & K=1   & 68.0/88.3 & 96.3/88.0 \\
    \multirow{3}{*}[+2.5ex]{w/o Multi-view}   & K=2   & 67.8/88.5 & 96.5/88.7 \\
          & K=4   & 69.1/88.9 & 96.8/88.9 \\
    \midrule
     & K=1   & 73.3/91.0 & 96.9/89.3 \\
    \multirow{3}{*}[+2.5ex]{w Multi-view}  & K=2   & 74.0/90.7 & 97.0/90.2 \\
          & K=4   & 70.7/89.7 & 96.9/88.9 \\
    \bottomrule
    \end{tabular}%
  \label{tab:addlabel}%
\end{table}%

\subsection{Ablation Study}
\subsubsection{Effect of multi-view features}
To validate the effectiveness of multi-view features, we conduct an ablation study. It can be seen that improvement brought by multi-view features is satisfactory in Table 5. Specifically, removing multi-view features enhancement results in a significant drop in performance, with classification metric I-AUROC decreasing by about 4.0\%, 4.4\%, 0.7\% and segmentation metric AUPRO decreasing by about  1.8\%, 1.6\%, 0.0\% in 1-shot, 2-shot and 4-shot setup respectively. We demonstrate that vision-language correlation ability of CLIP is further improved in anomaly detection benefiting from fused multi-view features. Projecting and rendering cloud into multi-view images not only mitigates modality gap but also enhances visual representation. The effectiveness of idea "Find anomalies from different views" is experimentally verified.

\subsubsection{Effect of multi-layer features}
Several methods which adapt CLIP for segmentation utilize the visual feature of single layer in CLIP~\cite{denseclip,maskclip} which is not sufficient to transfer high-quality visual representation of CLIP. We think that single stage of feature has limited representative ability and leads to loss of CLIP's prior knowledge. We combine $\{6, 9, 12\}$ layers of features as local visual representations for anomaly segmentation. To validate effectiveness of multi-layer aggregation, we conduct experiments on single layer $\{6\}$ and dual layers $\{6, 9\}$ and $\{9, 12\}$. As illustrated in Table 6, results of multi-layer $\{6, 9, 12\}$ surpass results of single last layer $\{6\}$ or dual layers $\{6, 9\}$ and $\{9, 12\}$ by a large margin comprehensively. Our experiments show that multi-layer features benefit the final performance in terms of anomaly classification and segmentation.

\begin{table}[t]
\caption{Exploration of contribution of multi-stage features in CLIP.}
  \centering
    \begin{tabular}{cccc}
    \toprule
       Layers  & K-shot & Classification  & Segmentation \\
    \midrule
     & K=1   & 68.3/88.4 & 95.9/86.1 \\
    \{6\}   & K=2   & 71.3/89.6 & 96.8/89.0 \\
          & K=4   & 65.7/87.3 & 96.9/89.5 \\
    \midrule
     & K=1   & 64.9/87.5 & 96.6/88.4 \\
    \{6, 9\}  & K=2   & 68.2/88.4 & 96.7/89.2 \\
          & K=4   & 69.9/89.2 & 97.0/89.8 \\
    \midrule
     & K=1   &71.0/89.6 & 96.4/88.0 \\
    \{9, 12\}  & K=2    & 70.8/90.0 & 96.5/87.9\\
          & K=4   & 67.1/87.7 & 96.7/88.8 \\
    \midrule
     & K=1   &  73.3/91.0 & 96.9/89.3 \\
    \{6, 9, 12 \}  & K=2   & 74.0/90.7 & 97.0/90.2 \\
          & K=4   & 70.7/89.7 & 96.9/88.9 \\
    \bottomrule
    \end{tabular}%
  \label{tab:addlabel}%
\end{table}%

\subsubsection{Impact of choice of views}
We project and render point cloud into 27 images totally. It is memory-costly and time-costly to choose all these images to process. Meanwhile, there is mass of redundant information among these images. In regard to selecting the multi-view images, we evaluate under four multi-view sets. The experiments presented in Table 7 show that results of set $\{5,9,14,19,27\}$ images achieve the best detection performance of 73.3\%, 74.0\%, 70.7\% I-AUROC and 89.3\%, 90.2\%, 88.9\% AUPRO for 1-shot, 2-shot and 4-shot. The results with fewer multi-view images $\{5,9,19,27\}$ are decreased by 5.9\%, 3.7\% for I-AUROC and 0.4\%, 1.6\% for AUPRO in 1-shot and 2-shot. On the contrary, adding image of new view will not improve performance and lead to increase of computation cost illustrated in third experiment of Table 7.

\begin{table}[t]
\caption{Exploration of influence of images with different views. }
  \centering
    \begin{tabular}{cccc}
    \toprule
       Views  & K-shot & Classification  & Segmentation \\
    \midrule
     & K=1   & 67.4/87.8 & 96.7/88.9 \\
    \multirow{3}{*}[+2.5ex]{\{5, 9, 19, 27\}}   & K=2   & 70.3/89.6 &96.6/88.6 \\
          & K=4   & 70.4/89.3 & 97.0/89.7 \\
    \midrule
     & K=1   & 71.1/90.1 &96.8/89.0 \\
    \multirow{3}{*}[+2.5ex]{\centering\{3, 7, 12, 17, 25\}}  & K=2   &72.8/90.2 &97.0/89.9 \\
          & K=4   & 72.0/90.3 & 96.5/88.0\\
    \midrule
     & K=1   & 72.3/90.5 & 96.4/88.4 \\
    \multirow{3}{*}[+2.5ex]{\centering\{5, 9, 14, 18, 19, 27\}}  & K=2   & 72.1/90.0 & 96.7/88.8 \\
          & K=4   & 73.0/90.4 & 96.3/87.8 \\
    \midrule
     & K=1   & 73.3/91.0 & 96.9/89.3 \\
    \multirow{3}{*}[+2.5ex]{\centering\{5, 9, 14, 19, 27\}}  & K=2   & 74.0/90.7 & 97.0/90.2 \\
          & K=4   &70.7/89.7 & 96.9/88.9 \\
    \bottomrule
    \end{tabular}%
  \label{tab:addlabel}%
\end{table}%

\subsection{Robustness to text prompts.}
Several prior works propose learnable prompts or well-designed compositional prompt ensemble to transfer CLIP for down-stream visual tasks~\cite{coop, clip-reid, winclip}. In our method, we mainly focus on adapting text representation extracted by CLIP instead of text prompts, which avoids struggling to design prompts artificially. Though we use compositional prompt ensemble (CPE)~\cite{winclip}, we still try to change text prompts and the results are shown in Figure 6. We change seven different sets of text prompts except for CPE. It is demonstrated that the classification and segmentation metrics vary slightly across different text prompts. Even if we use the simplest text prompts "a photo of a/the */damage [CLS]" (* means that we use no adjectives for normal samples), the classification metrics I-AUROC and AUPR even achieve 74.5\% and 91.4\% and the segmentation metrics AUPRO and P-AUROC reach 90.2\% and 97.2\% in 1-shot setup. Experiments demonstrate that given text prompts, text representations will be transferred to task-oriented space by adaptation. Robustness to text prompts of our method is  proved experimentally.

\subsection{Discuss of MVTEC-3D AD}
The appearance of MVTEC-3D AD dataset , a multimodal anomaly detection dataset including 3D point cloud and RGB images, is a milestone in the field of industrial anomaly detection. Anomalies are detected mainly based on 2D images in the past. However, nowadays in many real-world scenarios 2D RGB images are no longer sufficient because some geometry or structured anomalies are not obvious in 2D images but can be found in 3D point cloud easily as demonstrated in Figure 7 (a), which leads to terrible performance in some anomaly detection tasks. Meanwhile, some anomalies about color or texture information are more obvious in 2D images than which in 3D point cloud illustrated in Figure 7 (b). As we know, 3D point cloud is increasingly used in many real-world industrial scenarios. Combination of 2D images and 3D point cloud is a trend in the near future and they are both used to boost anomaly detection.

\begin{figure}[t]
\centering
\includegraphics[width=3.33in, keepaspectratio]{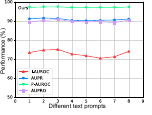}\\
\caption{Anomaly classification and segmentation performance with different text prompts. The first text prompts is compositional prompt ensemble used in our method. }
\end{figure}

\begin{figure}[t]
\centering
\includegraphics[width=3.33in, keepaspectratio]{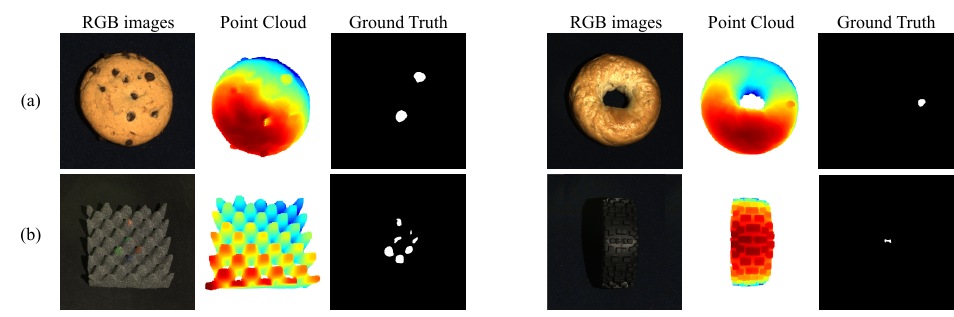}\\
\caption{Different anomalies are obvious in different modal data. (a) shows that surface or geometry anomalies are seen in 3D point cloud easily. While (b) shows color or texture anomalies are obvious in 2D images.}
\end{figure}

\section{Conclusion}
In this paper, we firstly introduce a novel 3D few-shot anomaly detection method CLIP3D-AD which effectively addresses both few-shot anomaly classification and segmentation and achieves the competitive performance without using memory banks and plenty of training samples on MVTEC-3D AD dataset.  We propose image and text adapters to adapt CLIP for anomaly classification and a coarse-to-fine image decoder to decode intermediate multi-layer features in CLIP image encoder for anomaly segmentation. Additionally, we project and render point cloud to multi-view images to utilize geometry information of point cloud and mitigate modal gap when using CLIP. A multi-view fusion module is proposed to fuse features of images with different views to enhance visual representations for further improving matching ability and feature discrimination. Experimental results demonstrate that our method is effective and ablation study further shows  the significant impact of proposed components in CLIP3D-AD.
%\bibliographystyle{ACM-Reference-Format}
%bibliography{reference}

% \appendix









\end{document}